%% file: main.tex
\pdfoutput=1

\documentclass[11pt]{article}

\usepackage[preprint]{acl}

\usepackage{times}
\usepackage{latexsym}
\usepackage{multirow}
\usepackage{colortbl}

\usepackage[T1]{fontenc}

\usepackage[utf8]{inputenc}

\usepackage{microtype}

\usepackage{inconsolata}

\usepackage{graphicx}

\usepackage{comment}

\newif{\ifMarginalComments}
\MarginalCommentstrue
\newif{\ifAdditionalMarginalComments}
\AdditionalMarginalCommentsfalse

\newcounter{ncomm}

\newcommand{\remove}[1]{}

\newcommand{\flowbench}{\textsc{Flow-Bench}}
\newcommand{\flowgen}{\textsc{Flow-Gen}}

\title{\flowbench: Towards Conversational Generation of Enterprise Workflows}

\author{
 \textbf{Evelyn Duesterwald},
 \textbf{Siyu Huo},
 \textbf{Vatche Isahagian},
 \textbf{K.R. Jayaram},
 \textbf{Ritesh Kumar},
\\
 \textbf{Vinod Muthusamy},
 \textbf{Punleuk Oum},
 \textbf{Debashish Saha},
 \textbf{Gegi Thomas},
 \textbf{Praveen Venkateswaran}
\\
\\
 \textbf{IBM Research AI},
\\
 \small{
 duester@us.ibm.com, siyu.huo@ibm.com,vatchei@ibm.com, jayaramkr@us.ibm.com, kumar.ritesh@ibm.com}
 \\ 
 \small{vmuthus@us.ibm.com, debashish.saha1@ibm.com, gegi@us.ibm.com, Praveen.Venkateswaran@ibm.com}
}

\begin{document}
\maketitle
\begin{abstract}

Business process automation (BPA) that leverages Large Language Models (LLMs) to convert natural language (NL) instructions into structured business process artifacts is becoming a hot research topic.
This paper makes two technical contributions -- (i) \flowbench, a high quality dataset of paired natural language instructions and structured business process definitions to
evaluate NL-based BPA tools, and support bourgeoning research in this area, and (ii) \flowgen,
our approach to utilize LLMs to translate natural language into an intermediate representation with Python syntax that facilitates final conversion into widely adopted business process definition languages,  such as BPMN and DMN. We bootstrap \flowbench\ by demonstrating how it can be used to evaluate the components of \flowgen\ across eight LLMs of varying sizes. We hope that \flowgen\ and \flowbench\ 
catalyze further research in BPA making it more accessible to novice and expert users.


\end{abstract}

\input{intro}

\input{dataset}
\input{approach}

\input{experiments}

\input{deployment}

\input{relatedwork}
\input{conclusion}

\bibliography{custom}

\appendix
\input{appendix}


\end{document}

%% file: intro.tex
\section{Introduction}

With many enterprises relying on BPA to standardize their work,
enhance their operational efficiency and reduce human error, BPA tools grew to a \$11.84B industry and are projected to grow to \$26B in 2028~\citep{marketwatch}.
In contemporary BPA tools, users use visual drag-and-drop interfaces and reusable templates to construct workflows, decision models, and document process logic, adhering to standard notations such as BPMN~\citep{grosskopf2009process, chinosi2012bpmn} and DMN~\citep{biard2015separation}.




However, even sophisticated low-code BPA platforms frequently necessitate intervention from software engineers to ensure robustness, handle intricate integrations, and implement customized logic not covered by generic templates. The complexity inherent in configuring detailed integration tasks and writing low-level transformation logic remains daunting for novice programmers and tedious even for seasoned developers. 


Recent efforts have explored leveraging natural language interfaces to simplify BPA authoring. LLMs have demonstrated substantial potential for automating code generation tasks by translating high-level user intents into executable artifacts. However, in our experience, even state-of-the-art LLMs are not effective at generating BPA workflows, both due to the lack of BPMN training data and the extensive boilerplate notations that need to be generated. Section \ref{sec:bpmnexample} showcases a flow and its corresponding BPMN code.

To the best of our knowledge, there is a lack of well-established benchmarks to evaluate NL-driven workflow generation. With this paper, we contribute \flowbench, a high quality dataset designed specifically to support research in natural language-driven business process automation, that consists of realistic utterances and their corresponding BPMN representations\footnote{\flowbench\; dataset can be accessed at https://github.com/IBM/Flow-Bench}. The availability of this dataset is intended to catalyze model improvements, benchmarking, and development of NLP techniques tailored to the BPA domain.

We also contribute \flowgen, an approach that leverages LLMs to  first translate natural language into an intermediate representation (IR) with Python syntax to precisely capture the logic of the intended business process. 
The Python IR provides several advantages. It takes advantage of the innate Python code generation capabilities of LLMs and it bridges the gap between unstructured natural language and formalized business process definition languages, facilitating easier verification and refinement of the generated logic. Subsequently, our approach translates this IR into specific target process definition languages, such as BPMN-compliant XML or DMN decision tables, ensuring compatibility with existing BPA solutions and tools. Further, the use of an IR makes it easy to catch errors early and to support multiple target BPA languages.  We include the Python IR in the \flowbench\ dataset, along with the BPMN.

%% file: dataset.tex
\section{\flowbench\ dataset}
\label{sec:Dataset}

Workflows in \flowbench\ consist of sequences of API invocations, including conditionals and loops, similar to those found in commercial workflow automation platforms like IBM App Connect~\cite{ibmappconnecttemplates} and Zapier~\cite{zapierapps}. In addition to API calls, workflows may incorporate manual interventions, termed \textit{user tasks}. These user tasks typically involve steps within a business process that require human action, such as managerial approvals, and thus, do not have corresponding APIs.

To construct \flowbench\ we initially sourced realistic business workflows from pre-existing templates provided by commercial workflow automation platforms (specifically IBM App Connect and Zapier). These templates cover common enterprise use cases, including support ticket creation, task management, and marketing campaign automation.

We carried out three high level steps to arrive at the final \flowbench\ dataset: (1) quality control, (2) manual labeling, and (3) data augmentation.

\paragraph{Quality Control:} First, we collected and manually examined workflows sourced from the automation platforms. We discarded or truncated workflows that were overly complex with multiple levels of nested conditions, in order to start with relatively small workflows that a user can reasonably describe in a few sentences.
We removed event triggers from workflows and discarded workflows that consisted of only a single API call. We also discarded workflows that involved APIs without publicly available OpenAPI spec.

\paragraph{Manual Labeling:} We manually added or corrected user utterance to workflows. In some cases, we rephrased existing workflow descriptions to reflect an active user command to constructing the workflow as opposed to a passive description of the workflow. The workflows often only had a proprietary representation in the respective automation tool, so we manually crafted Python IR  snippets and generated the corresponding BPMN specification. We also retrieved the OpenAPI specs for all activities in the workflows and established a common naming convention for API-based worfklow activities. In addition, we added clear descriptions for each API, tweaking the descriptions in the OpenAPI specs (if available).

\paragraph{Data Augmentation:} Once we had a set of high quality samples, we expanded the dataset in two ways. First we incorporated user tasks by either adding a new activity in the workflow, or by removing the corresponding API from the catalog. Then we added samples to reflect how users may incrementally build a workflow, by adding, deleting, or replacing activities in the workflow. For each new sample, we defined the workflow before and after the edit, and the corresponding user utterance. To mimic actual software development the incremental edits may apply anywhere in the current workflow, not always editing left to right.


The final complete workflows in \flowbench\ are generated through incremental build steps categorized as \textit{add}, \textit{delete}, or \textit{replace}. 

Build steps in \flowbench\ are kept as self-contained tests by including three elements: (1) \textit{Prior Sequence}, representing the current state of the workflow before applying changes\footnote{This may be empty in the initial generation step.}; (2) \textit{Utterance}, describing the modification to be performed in natural language (e.g, "Retrieve all issues from the Jira board."); and (3) \textit{Expected Sequence}, indicating the resulting workflow after implementing the command specified by the utterance. 


\flowbench\ comprises 101 incremental build step tests structured according to this methodology. Figure~\ref{fig:flowbench_test} illustrates an example test case from \flowbench. Each build step is uniquely identified and provides a self-contained test scenario, including explicit BPMN representations of both the \textit{Prior Sequence} and \textit{Expected Sequence}. Additionally, each test is annotated with metadata describing the build step type (\texttt{add}, \texttt{delete}, \texttt{replace}), the control flow structure (\texttt{linear} or \texttt{conditional}, where \texttt{conditional} encompasses both if-statements and loops), and the inclusion of user task steps (denoted by \texttt{user\_task}). BPMN representation of Figure~\ref{fig:flowbench_test} is shown in \ref{sec:app:bpmn97}


To ensure compact and clear representations of prior and expected workflows, \flowbench\ adopts a constrained subset of Python syntax. This subset includes assignment statements, conditional statements (if-statements), loops (for and while), and function calls. These representations are explicitly provided within the \textit{Prior Sequence} and \textit{Expected Sequence} elements of each test.

\begin{figure}
{\tiny

\begin{verbatim}
_metadata:
  tags:
    - conditional_update
  uid: 97
input:
  utterance: |-
    Instead of retrieving all the issues 
    just create a new issue in each repo
  prior_sequence:
    - |-
      repositories = GitHub_Repository__3_0_0__retrievewithwhere_Repository()
      for repo in repositories:
        new_issue = GitHub_Issue__3_0_0__retrievewithwhere_Issue()
  prior_context: []
  bpmn:
    $ref: "context/uid_97_context.bpmn"
expected_output:
  sequence:
    - |-
      repositories = GitHub_Repository__3_0_0__retrievewithwhere_Repository()
      for repo in repositories:
        updated_issue = GitHub_Issue__3_0_0__create_Issue()
  bpmn:
    $ref: "output/uid_97_output.bpmn"
\end{verbatim}
}
\caption{\footnotesize Example of \flowbench\ test case}
\label{fig:flowbench_test}
\end{figure}


Generating accurate pythonic function calls in a \flowbench\ test by an LLM requires knowledge of existing APIs and descriptions. Thus, we also provide a separate file containing a list of APIs along with their descriptions. An example of the API and the description is shown below.

{\scriptsize
\begin{verbatim}
      {
        "id": "Jira_Issue__2_0_0__retrievewithwhere_Issue",
        "description": "Retrieve all Jira issues"
      }
\end{verbatim}
}

%% file: approach.tex
\section{\flowgen\ }
\label{sec:approach}

\begin{figure*}[htp!]
    \centering
    \captionsetup{font=footnotesize}
    \includegraphics[width=0.90\linewidth]{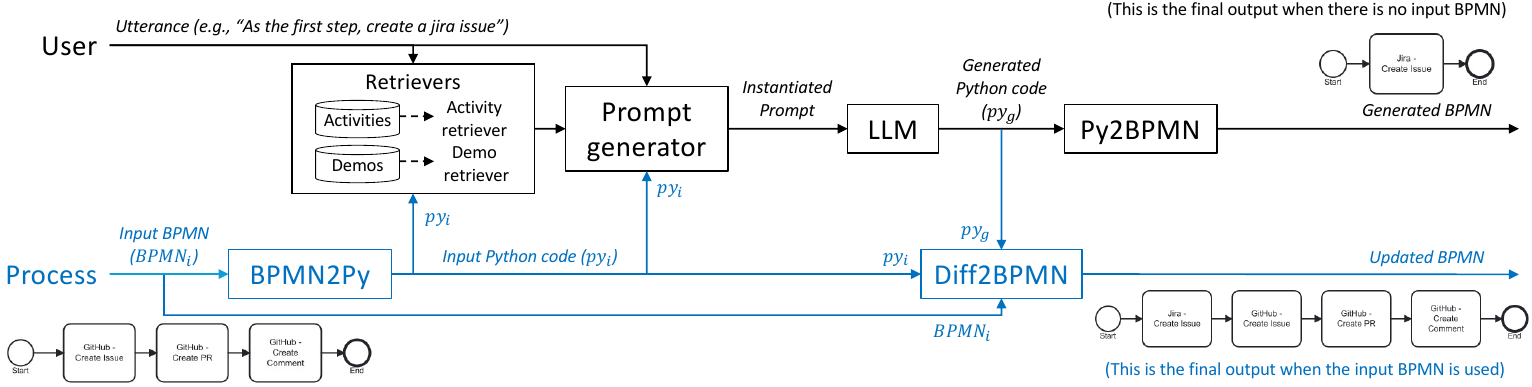}
    \caption{\flowgen\ overview. The top part (in black) depicts the steps to generate a new workflow based on a user utterance. The bottom part (in blue) are the additional steps to update an existing workflow based on an utterance.}
    \label{fig:approach-overview}
\end{figure*}

This section presents \flowgen, an approach that applies pre-trained LLMs to solve the workflow generation tasks in the \flowbench\ dataset.

\subsection{Python Intermediate Representation}

Our initial observations suggest that while some LLMs can generate BPMN directly, these models have not been extensively trained or evaluated specifically on BPMN data. In contrast, many pre-trained LLMs have been extensively trained on Python code and have demonstrated significant proficiency in generating Python scripts accurately~\cite{codejudge}.

Moreover, BPMN representations are inherently verbose, which introduces complexity and increases the likelihood of syntactic and semantic errors when generated by LLMs. Longer BPMN outputs also increase computational cost and slow down generation, negatively impacting interactive user experiences. To illustrate this, consider the straightforward linear BPMN flow depicted in the bottom left of Figure~\ref{fig:approach-overview}. Its BPMN representation requires 3,151 characters, whereas the equivalent logic can be succinctly expressed in only 148 characters of Python code:

{\scriptsize
\begin{verbatim}
        issue = GitHub_Issue__3_0_0__create_Issue()
        pr = GitHub_Pullrequest__3_0_0__create_Pullrequest()
        comment = GitHub_Comment__3_0_0__create_Comment()
\end{verbatim}}

In \flowbench, the BPMN representation is on average 25 times longer than the Python equivalent. While there are BPMN-specific concepts such as swimlanes and roles that do not translate directly into Python, these are out of scope of the \flowbench\ dataset. \flowgen\ also accommodates the concept of user tasks—activities performed by humans that are not linked to predefined workflows or APIs. These tasks represent ad-hoc activities specified by the user.

\subsection{Initial Flow Generation}

Consider the scenario where a user wants to create a new workflow based solely on an NL description. This initial flow generation process, outlined in the upper part of Figure~\ref{fig:approach-overview}, involves several steps.

Initially, the user's NL utterance is analyzed to retrieve a relevant subset of predefined activities. These activities typically correspond to APIs, decision rules, or other processes from a comprehensive activity catalog. Including this subset in the LLM prompt is essential, especially when the entire catalog cannot fit within the LLM's context window. The different approaches to retrieving relevant activities are detailed in Section~\ref{sec:activities-retrievers}.

Concurrently, the utterance is used to select the most relevant demonstrations from the dataset. Each demonstration comprises an NL utterance paired with a Python code snippet. These examples, provided as few-shot demonstrations in the LLM prompt, guide the LLM to generate accurate Python code snippets, illustrating correct invocation patterns of predefined activities as Python functions. Section~\ref{sec:demonstrations-retrievers} further elaborates on the methods evaluated for demonstration retrieval.

Next, an LLM generates a Python code snippet based on a dynamically assembled prompt that includes the user's NL utterance, retrieved activity descriptions, and selected few-shot demonstrations. This generated snippet captures the workflow described by the user, incorporating user tasks where necessary for activities not in the catalog.

Finally, the deterministic \textsc{Py2BPMN} module converts the generated Python code into standard BPMN, completing the translation from NL to executable workflow definition.

\subsection{Incremental Flow Updates}

Let us now consider the case where there is already an existing workflow, and the user issues an utterance to incrementally edit the workflow. The bottom portion of Figure~\ref{fig:approach-overview} show the additional steps in \flowgen\ to support this case.

First, the original BPMN workflow (${BPMN}_i$) is transformed into a Python code snippet (${py}_i$). This is done using deterministic code in the \textsc{BPMN2Py} module.

The code ${py}_i$ is used as additional information for the retrievers. For example, the activity retriever should select not only activities mentioned in the utterance, but also those referenced in the input workflow. Similarly, if ${py}_i$ contains conditional or looping constructs, the demonstration retriever will more likely select few-shot samples that include such constructs.
The code snippet ${py}_i$ is coupled with the user query to serve as input to the LLM code generation step.

The \textsc{Diff2BPMN} module computes the difference between the input (${py}_i$) and generated (${py}_g$) Python and internally generates a set of update operations. These update operations are applied to the input workflow (${BPMN}_i$) to arrive at the final updated BPMN workflow.


\subsection{Activity Retrievers} \label{sec:activities-retrievers}

Given the user's NL utterance, the generated code is expected to reference activities from the provided catalog (grounding) and avoid hallucinations. However, as catalogs may contain thousands of activities, it becomes infeasible to include the entire catalog within the limited context window of the LLM. Thus we need to retrieve and include only the top-k most relevant subset. We outline three types of activity retrievers.

\textbf{ED\_Retriever}: Compares the user utterance against the description of each activity from the catalog to quantify how dissimilar (or similar) the two are based on edit distance. Given that edit distance computation only compares the raw strings without incorporating any semantic meaning, the performance of this retriever is limited in scope.

\textbf{Embeddings\_Retriever}: A Bi-Encoder based retrieval that generates the embedding vectors for the user utterance and all the activities, followed by computing the cosine similarity between each pair. Embeddings capture the semantic meaning and thus, boost the performance significantly as compared to the Edit Distance based approaches. We used the \textit{all-MiniLM-L6-v2}\footnote{https://huggingface.co/sentence-transformers/all-MiniLM-L6-v2}  model to generate the embeddings and ChromaDB to store, index, and retrieve top-K activities. Since the catalog is relatively stable, the embeddings can be generated once and stored to improve runtime latency. 

\textbf{Activities\_Search}: This retriever works like Embeddings\_Retriever, but we use a custom model fine-tuned to generate better embeddings for the activity retrieval task.



\subsection{Demonstration Retrievers}
\label{sec:demonstrations-retrievers}

Demonstrations refer to the few shot in-context examples that are incorporated in the prompt. We explored two retrieval approaches.
 
 \textbf{TopKRetriever}: A Bi-Encoder based retrieval similar to Embeddings\_Retriever. 
 
\textbf{CE\_Retriever}: A cross-encoder based retrieval, where two strings are passed simultaneously to the model that outputs similarity score ranging between 0 and 1. 
Cross-encoder based retrieval is more accurate since the model is trained on a large dataset to generate the similarity score. Since the user utterance is only available at runtime, the similarity computation against the demos can only be performed at runtime which introduces latency. To reduce the latency, we shortlist the demonstration catalog based on the provided context before passing it to cross-encoder. For example, if the user is updating an existing workflow, only the demos containing a prior sequence are selected.
We use the \textit{stsb-distilroberta-base}\footnote{https://huggingface.co/cross-encoder/stsb-distilroberta-base} model as the cross-encoder in our experiments.

%% file: experiments.tex
\section{Evaluation}
\label{sec:evaluation}
In this section, we evaluate \flowgen\ on the \flowbench\ dataset. 
We begin by providing an evaluation of the retrievers.
All experiments were conducted over the 101 \flowbench\ test cases.


\subsection{Activity Retrievers
} \label{sec:shortlister}


\begin{table}[htbp]
\captionsetup{font=footnotesize}
\resizebox{\columnwidth}{!}{%
\begin{tabular}{|lcccc|}
\hline
{ \textbf{Retriever}} & \textbf{TopK} & \textbf{Activities Recall}& \textbf{Exact Match} & \textbf{Hallucination Rate} \\ 
\hline
{ } & 10 & 0.7327 & 0.495 & \textbf{0.0469} \\
{ } & 50 & 0.7913 & 0.604 & 0.0621 \\
\multirow{-3}{*}{{ \textbf{ED\_Retriever}}} & {100} & \textbf{0.8086} & \textbf{0.6139} & 0.0586 \\ 
\hline
{ } & 10 & 0.9307 & 0.6733 & 0.0207 \\
{ } & {50} & 0.9794 & \textbf{0.7525} & \textbf{0.0205} \\
\multirow{-3}{6.0em}{{ \textbf{Embeddings\_Retriever}}} & 100 & \textbf{0.9851} & 0.7129 & 0.0236 \\ 
\hline
{ } & 10 & 0.9703 & 0.6931 & 0.0069 \\
{ } & {50} & \textbf{0.9926} &\textbf{0.7723} & \textbf{0.0102} \\
\multirow{-3}{*}{{ \textbf{Activities\_Search}}} & 100 & \textbf{0.9926} & 0.7525 & \textbf{0.0102} \\ 
\hline
\end{tabular}
}
\caption{Evaluation of different retrievers with different TopK.}
\label{tab:apiShortLister}
\end{table}

Table~\ref{tab:apiShortLister} summarizes the results of the three activity retrievers: ED\_Retriever, Embeddings\_Retriever, and Activities\_Search.
\textbf{TopK} refers to the number of retrieved activities. \textbf{Activities Recall} is computed based on the overlap between the retrieved activities and those in the ground truth.
\textbf{Exact Match} highlights the accuracy of the generated IR to the ground truth syntactically and semantically.
\textbf{Hallucination Rate} computes the fraction of activities in the generated workflows that are not in the catalog. 

Table~\ref{tab:apiShortLister} shows that Activities\_Search with TopK=50 has the best recall and best exact match score with the least hallucination rate, followed by Embeddings\_Retriever. This validates that embedding similarity is more effective than edit distance. We also see that larger TopK improves recall but reduces the exact match since it increases LLM's probability of selecting the incorrect Activity. 



\begin{table}[htbp]
\captionsetup{font=footnotesize}
\resizebox{\columnwidth}{!}{%
\begin{tabular}{|llll|}
\hline
\textbf{Model} & \textbf{DemoSelector} & \textbf{TopK} & \textbf{Exact Match} \\ \hline
 &  & 2 & 0.5842 \\
 &  & 3 & 0.5941 \\
 &  & \textbf{5} & \textbf{0.6734} \\
 & \multirow{-4}{*}{TopKRetriever} & 7 & 0.6634 \\ \cline{2-4} 
 &  & 2 & 0.5842 \\
 &  & 3 & 0.6436 \\
 &  & \textbf{5} & \textbf{0.6931} \\
\multirow{-8}{*}{\textbf{Granite-20b-code-instruct-v2}} & \multirow{-4}{*}{CE\_Retriever} & 7 & 0.6733 \\ \hline
 &  & 2 & 0.7029 \\
 &  & 3 & 0.7131 \\ 
 &  & \textbf{5} & \textbf{0.7228} \\
 & \multirow{-4}{*}{TopKRetriever} & \textbf{7} & \textbf{0.7228} \\ \cline{2-4} 
 &  & 2 & 0.703 \\
 &  & 3 & 0.7228 \\
 &  & \textbf{5} & \textbf{0.7624} \\
\multirow{-8}{*}{\textbf{codellama-34b-instruct-hf}} & \multirow{-4}{*}{CE\_Retriever} & \textbf{7} & \textbf{0.7624} \\ \hline
\end{tabular}%
}
\caption{Evaluation of demonstration retrievers with different number of demos (TopK) while using Activities\_Search with TopK=50 for Activities selection.}
\label{tab:demoSelector}
\end{table}

\subsection{Demonstration Retriever
}
\label{sec:demoselector}
Table \ref{tab:demoSelector} compares \textit{\textbf{TopKRetriever}} and  \textit{\textbf{CE\_Retriever}} for different values of TopK
using Granite-20b-code-instruct-v2 and codellama-34b-instruct-hf models.
TopK here refers to the number of demonstrations not activities.

The cross-encoder based retriever boosts the exact match by 4 points irrespective of model choice. Increasing the number of demonstrations retrieved beyond five degrades the overall performance. For the remainder of these experiments we consider retrieving five demonstrations using CE\_Retriever.



\begin{table}[htbp]
\centering
\small
\captionsetup{font=footnotesize}
\resizebox{\linewidth}{!}{  
\begin{tabular}{|l|l|l|l|}
\hline
\textbf{Model} & \textbf{Activities Domain} & \textbf{Exact Match} & \textbf{Syntax F1}\\ \hline
 & in-domain & 0.66 & 0.87\\
\multirow{-2}{*}{\textbf{mixtral-8x7b-instruct-v01}} & cross-domain & 0.63 & 0.85\\ \hline
 & in-domain & 0.59 & 0.86\\
\multirow{-2}{*}{\textbf{granite-8b-code-instruct}} & cross-domain & 0.60 & 0.84 \\ \hline
 & in-domain & 0.19 & 0.57\\
\multirow{-2}{*}{\textbf{llama-3-1-8b-instruct}} & cross-domain & 0.19 & 0.56 \\ \hline
 & in-domain & 0.67 & 0.91\\
\multirow{-2}{*}{\textbf{Granite-20b-code-instruct-v2}} & cross-domain & 0.59 & 0.89\\ \hline
 & in-domain & 0.76 & 0.93\\
\multirow{-2}{*}{\textbf{Codellama-34b-instruct-hf}} & cross-domain & 0.72 & 0.91\\ \hline
 & in-domain & 0.53 & 0.74\\
\multirow{-2}{*}{\textbf{llama-3-3-70b-instruct}} & cross-domain & 0.49 & 0.71\\ \hline
 & \textbf{in-domain} & \textbf{0.83} & \textbf{0.90} \\
\multirow{-2}{*}{\textbf{Mistral-large}} & \textbf{cross-domain} & \textbf{0.79} & \textbf{0.86} \\ \hline
 & in-domain & 0.60 & 0.82 \\
\multirow{-2}{*}{\textbf{llama-3-405b-instruct}} & cross-domain & 0.55 & 0.80 \\ \hline
\end{tabular} 
}
\caption{Evaluation of different models using Activities\_Search (TopK=50) as and CE\_Retriever(TopK=5) as Activities and demos retrievers respectively.}
\label{tab:evaluation}
\end{table}

\subsection{Overall Evaluation}
In Table~\ref{tab:evaluation}, we provide an extensive evaluation of \flowgen\. We use Activities\_Search as the activity retriever with TopK=50 and CE\_Retriever as the demonstration retriever with TopK=5. Table \ref{tab:evaluation} compares the performance of several models varying in size. Recall that \textbf{Exact Match} highlights the accuracy of the generated IR to the ground truth syntactically and semantically, and \textbf{Activities Recall} is the overlap between the retrieved activities and those in the ground truth. \textbf{Syntax F1} evaluates correctness of the generated IR code syntactically.

To evaluate the impact of interference between the activities catalog and activities present in the demonstrations we provide both \textbf{cross-domain} and \textbf{in-domain} evaluations. By \textbf{cross-domain} we make sure that demos are selected such that the activities present in the ground-truth are not used by any of the selected demonstrations and for \textbf{in-domain}, activities present in ground-truth may be present in selected demos.

Mistral-large model preformed best with exact match of  0.83 and 0.79
for both in-domain and cross-domain scenarios respectively. High Syntax F1 highlights the ability of Mistral-large to generate syntactically correct Python IRs. The llama-3-1-8b-instruct small model performs the worst.



%% file: deployment.tex
\section{Deployment}
\label{sec:deployment}

\begin{figure}[htp]
    \centering
    \captionsetup{font=footnotesize}
    \includegraphics[width=0.87\linewidth]{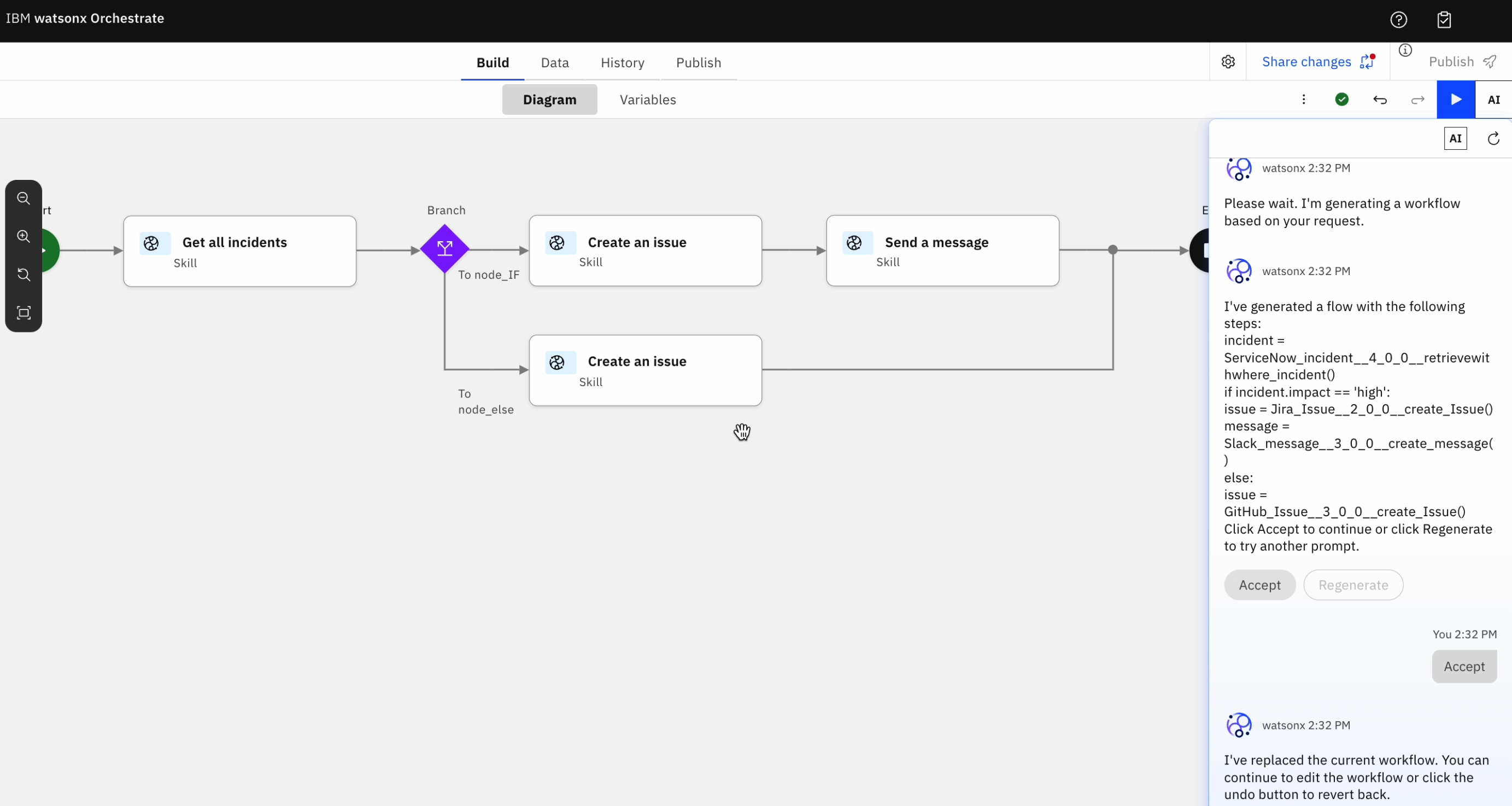}
    \caption{Deployment in WxO production environment}
    \label{fig:wxo}
\end{figure}

Our approach has been deployed as a technical preview (c.f. Figure \ref{fig:wxo}) as part of the Unified Automation Builder (UAB) of IBM’s Watsonx Orchestrate (\citeyear{ibmwatsonorchestrate}). UAB provides an intuitive graphical interface for the creation, evaluation and deployment of automation flows. The UAB tooling is deployed as a scalable cloud solution, with numerous containers deployed in a 
cluster. Our approach has been to deploy \flowgen\ as a first-class component, in the cluster, enabling secure access to the API catalog, as well as LLM inference capabilities available from Watsonx.ai. Additionally, Watsonx Assistant is leveraged as the user-facing interface, where utterances are input by the user, and routed to \flowgen\ via internal proxy services which facilitate the returning responses.

%% file: relatedwork.tex
\section{Related Work}
\label{sec:relatedwork}

Leveraging LLMs to automate the creation and improvement of BPM flows is an active area of research. AutoFlow~\cite{Li2024AutoFlowAW} is a framework that automatically generates workflows enabling agents to tackle complex tasks. It adopts the CoRE language~\cite{xu2024core} for workflow representation, requiring fine-tuning of LLMs to master the specific grammar and workflow generation protocols associated with CoRE. However, this fine-tuning requirement limits flexibility, preventing the integration of off-the-shelf LLMs.

Agentic Process Automation (APA)~\cite{Ye2023ProAgentFR} formulates workflow creation as a Python code-generation task, where actions within the workflows are represented by Python function calls executed by specialized agents. Nonetheless, APA does not ground the APIs explicitly within the business process, creating potential for hallucination where the LLM might reference incorrect or nonexistent APIs. Our research avoids hallucination by embedding API grounding directly within the LLM prompts, thus enabling the LLM to accurately interpret, select, and utilize APIs as practical tools for workflow generation~\cite{Liu2024ToolACEWT,berkeley-function-calling-leaderboard,Qin2023ToolLLMFL}. This also facilitates the LLM's comprehension APIs, resulting in Python-generated workflows that reflect the intended business process control flow.

Recently, \cite{Fan2024WorkflowLLMEW} also used a Pythonic IR for workflow processes and collected grounded APIs specifically for workflow construction. Their workflow generation strategy depends on training specialized data annotators based on data collected, which is constrained by the domains of data collection (Apple Shortcuts and RoutineHub), limiting its broader applicability handling out-of-domain APIs and queries. In contrast, our method leverages in-context learning combined with API retrieval techniques, both of which inherently support greater flexibility and ease of generalization across diverse domains. In addition, our approach supports multi-turn conversational workflow construction.

%% file: conclusion.tex
\section{Conclusions}
\label{sec:conclusion}

In this paper, our contributions include both the \flowbench\ dataset and the \flowgen\ methodology, aiming to significantly lower barriers for both expert and citizen developers to construct 
automated business processes.
\flowbench\ is a novel dataset specifically curated to facilitate research on automating workflows using LLMs. \flowgen\ is a technique that translates natural language instructions into structured BPMN artifacts, leveraging an IR. By grounding workflow construction in contextually relevant API documentation and utilizing in-context learning, we address common issues such as hallucination and limited domain adaptability present in previous methods. Future work includes expanding dataset coverage, further refining API grounding methods, and evaluating our proposed methods in more complex business scenarios.

%% file: appendix.tex
\newpage
\section{Appendix}
\label{sec:appendix}

\subsection{BPMN diagram and code}
\label{sec:bpmnexample}
\begin{figure} [htp]
    \centering
    \includegraphics[width=1\linewidth]{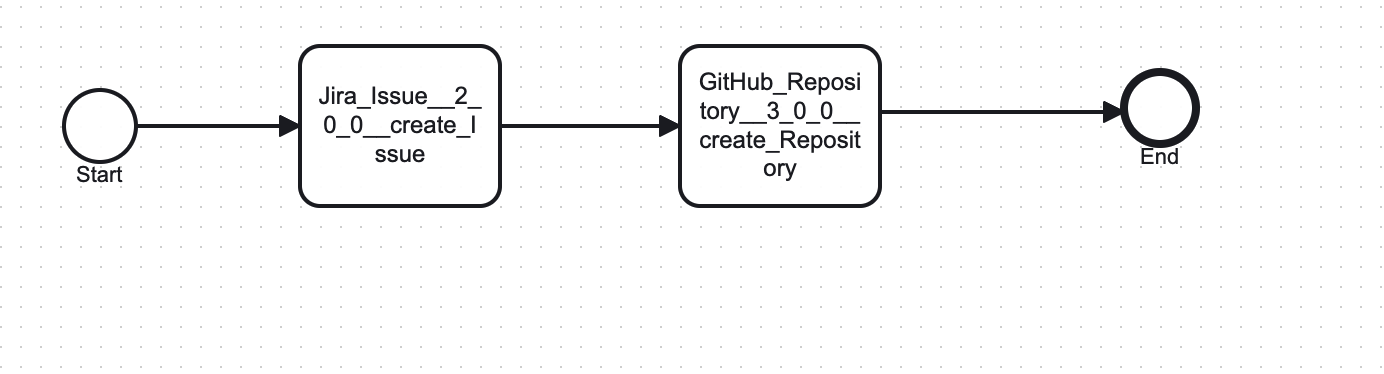}
    \caption{Simple flow example}
    \label{fig:flow1}
    \vspace{-0.2in}
\end{figure}

Figure \ref{fig:flow1} shows an example of a flow expressed as a BPMN diagram along with the corresponding BPMN code shown in Figure \ref{fig:flow1_bpmn}.

\begin{figure} [hp]
{\tiny
\begin{verbatim}
<?xml version="1.0" encoding="UTF-8"?>
<definitions xmlns="http://www.omg.org/spec/BPMN/20100524/MODEL" 
xmlns:bpmndi="http://www.omg.org/spec/BPMN/20100524/DI" 
xmlns:dc="http://www.omg.org/spec/DD/20100524/DC" 
xmlns:di="http://www.omg.org/spec/DD/20100524/DI" exporter="bpmn-js
(https://demo.bpmn.io)" exporterVersion="18.3.1">
  <process id="Process_1" isExecutable="false">
    <startEvent id="startEvent_1" name="Start" />
    <task id="task_2" name="Jira_Issue__2_0_0__create_Issue" />
    <sequenceFlow id="flow_startEvent_1_task_2" 
    sourceRef="startEvent_1" targetRef="task_2" />
    <task id="task_3" name="GitHub_Repository__3_0_0__create_Repository" />
    <sequenceFlow id="flow_task_2_task_3" 
    sourceRef="task_2" targetRef="task_3" />
    <endEvent id="endEvent_4" name="End" />
    <sequenceFlow id="flow_task_3_endEvent_4" 
    sourceRef="task_3" targetRef="endEvent_4" />
  </process>
  <bpmndi:BPMNDiagram id="BPMNDiagram_1">
    <bpmndi:BPMNPlane id="BPMNPlane_1" bpmnElement="Process_1">
      <bpmndi:BPMNShape id="BPMNShape_task_3" bpmnElement="task_3">
        <dc:Bounds x="460" y="80" width="100" height="80" />
      </bpmndi:BPMNShape>
      <bpmndi:BPMNShape id="BPMNShape_endEvent_4" 
      bpmnElement="endEvent_4">
        <dc:Bounds x="682" y="93" width="36" height="36" />
        <bpmndi:BPMNLabel>
          <dc:Bounds x="690" y="129" width="20" height="14" />
        </bpmndi:BPMNLabel>
      </bpmndi:BPMNShape>
      <bpmndi:BPMNShape id="BPMNShape_task_2" bpmnElement="task_2">
        <dc:Bounds x="270" y="80" width="100" height="80" />
      </bpmndi:BPMNShape>
      <bpmndi:BPMNShape id="BPMNShape_startEvent_1" 
      bpmnElement="startEvent_1">
        <dc:Bounds x="152" y="102" width="36" height="36" />
        <bpmndi:BPMNLabel>
          <dc:Bounds x="158" y="138" width="24" height="14" />
        </bpmndi:BPMNLabel>
      </bpmndi:BPMNShape>
      <bpmndi:BPMNEdge id="BPMNEdge_flow_startEvent_1_task_2" 
      bpmnElement="flow_startEvent_1_task_2">
        <di:waypoint x="188" y="120" />
        <di:waypoint x="270" y="120" />
      </bpmndi:BPMNEdge>
      <bpmndi:BPMNEdge id="BPMNEdge_flow_task_2_task_3" 
      bpmnElement="flow_task_2_task_3">
        <di:waypoint x="370" y="120" />
        <di:waypoint x="460" y="120" />
      </bpmndi:BPMNEdge>
      <bpmndi:BPMNEdge id="BPMNEdge_flow_task_3_endEvent_4" 
      bpmnElement="flow_task_3_endEvent_4">
        <di:waypoint x="560" y="113" />
        <di:waypoint x="682" y="113" />
      </bpmndi:BPMNEdge>
    </bpmndi:BPMNPlane>
  </bpmndi:BPMNDiagram>
</definitions>

\end{verbatim}
}
\caption{BPMN code corresponding to example in Figure \ref{fig:flow1}}
\label{fig:flow1_bpmn}
\end{figure}

\newpage
\subsection{BPMN Representation of \flowbench\ test case}
\label{sec:app:bpmn97}
Figures \ref{fig:prior_sequence} and \ref{fig:final_output} show the BPMN code for \flowbench test case shown in Figure \ref{fig:flowbench_test}.

\begin{figure*}
{\tiny
\begin{verbatim}
<?xml version="1.0" encoding="UTF-8"?>
<bpmn:definitions xmlns:xsi="http://www.w3.org/2001/XMLSchema-instance" xmlns:bpmn="http://www.omg.org/spec/BPMN/20100524/MODEL" 
xmlns:bpmndi="http://www.omg.org/spec/BPMN/20100524/DI" xmlns:dc="http://www.omg.org/spec/DD/20100524/DC" 
xmlns:di="http://www.omg.org/spec/DD/20100524/DI" exporter="Camunda Modeler" 
exporterVersion="5.32.0">
  <bpmn:process id="Process_1j6betq" isExecutable="false">
    <bpmn:startEvent id="StartEvent_1twgfyv">
      <bpmn:outgoing>Flow_040uk43</bpmn:outgoing>
    </bpmn:startEvent>
    <bpmn:subProcess id="Activity_0n3dkn6">
      <bpmn:incoming>Flow_0fmvfja</bpmn:incoming>
      <bpmn:outgoing>Flow_03o4pmp</bpmn:outgoing>
      <bpmn:multiInstanceLoopCharacteristics isSequential="true" />
      <bpmn:startEvent id="Event_1g6k28n">
        <bpmn:outgoing>Flow_0ez4w93</bpmn:outgoing>
      </bpmn:startEvent>
      <bpmn:endEvent id="Event_0lbdydr">
        <bpmn:incoming>Flow_05t21yg</bpmn:incoming>
      </bpmn:endEvent>
      <bpmn:sequenceFlow id="Flow_0ez4w93" sourceRef="Event_1g6k28n" targetRef="Activity_0sj4qjl" />
      <bpmn:task id="Activity_0sj4qjl"      name="GitHub_Issue__3_0_0__retrievewithwhere_Issue">
        <bpmn:incoming>Flow_0ez4w93</bpmn:incoming>
        <bpmn:outgoing>Flow_05t21yg</bpmn:outgoing>
      </bpmn:task>
      <bpmn:sequenceFlow id="Flow_05t21yg" sourceRef="Activity_0sj4qjl" targetRef="Event_0lbdydr" />
    </bpmn:subProcess>
    <bpmn:endEvent id="Event_1ycwwda">
      <bpmn:incoming>Flow_03o4pmp</bpmn:incoming>
    </bpmn:endEvent>
    <bpmn:sequenceFlow id="Flow_03o4pmp" sourceRef="Activity_0n3dkn6" targetRef="Event_1ycwwda" />
    <bpmn:task id="Activity_0cwpd7f" name="GitHub_Repository__3_0_0__retrievewithwhere_Repository">
      <bpmn:incoming>Flow_040uk43</bpmn:incoming>
      <bpmn:outgoing>Flow_0fmvfja</bpmn:outgoing>
    </bpmn:task>
    <bpmn:sequenceFlow id="Flow_040uk43" sourceRef="StartEvent_1twgfyv" targetRef="Activity_0cwpd7f" />
    <bpmn:sequenceFlow id="Flow_0fmvfja" sourceRef="Activity_0cwpd7f" targetRef="Activity_0n3dkn6" />
    <bpmn:textAnnotation id="TextAnnotation_1q9vfnx">
      <bpmn:text>for repo in repositories</bpmn:text>
    </bpmn:textAnnotation>
    <bpmn:association id="Association_0c0ii8c" associationDirection="None" sourceRef="Activity_0n3dkn6" targetRef="TextAnnotation_1q9vfnx" />
  </bpmn:process>
  <bpmndi:BPMNDiagram id="BPMNDiagram_1">
    <bpmndi:BPMNPlane id="BPMNPlane_1" bpmnElement="Process_1j6betq">
      <bpmndi:BPMNShape id="_BPMNShape_StartEvent_2" bpmnElement="StartEvent_1twgfyv">
        <dc:Bounds x="152" y="177" width="36" height="36" />
      </bpmndi:BPMNShape>
      <bpmndi:BPMNShape id="Activity_0cwpd7f_di" bpmnElement="Activity_0cwpd7f">
        <dc:Bounds x="250" y="155" width="100" height="80" />
        <bpmndi:BPMNLabel />
      </bpmndi:BPMNShape>
      <bpmndi:BPMNShape id="Event_1ycwwda_di" bpmnElement="Event_1ycwwda">
        <dc:Bounds x="1102" y="177" width="36" height="36" />
      </bpmndi:BPMNShape>
      <bpmndi:BPMNShape id="Activity_0dishkm_di" bpmnElement="Activity_0n3dkn6" isExpanded="true">
        <dc:Bounds x="440" y="130" width="500" height="150" />
      </bpmndi:BPMNShape>
      <bpmndi:BPMNShape id="Event_1g6k28n_di" bpmnElement="Event_1g6k28n">
        <dc:Bounds x="472" y="182" width="36" height="36" />
      </bpmndi:BPMNShape>
      <bpmndi:BPMNShape id="Activity_0sj4qjl_di" bpmnElement="Activity_0sj4qjl">
        <dc:Bounds x="650" y="160" width="100" height="80" />
        <bpmndi:BPMNLabel />
      </bpmndi:BPMNShape>
      <bpmndi:BPMNShape id="Event_0lbdydr_di" bpmnElement="Event_0lbdydr">
        <dc:Bounds x="842" y="182" width="36" height="36" />
      </bpmndi:BPMNShape>
      <bpmndi:BPMNEdge id="Flow_05t21yg_di" bpmnElement="Flow_05t21yg">
        <di:waypoint x="750" y="200" />
        <di:waypoint x="842" y="200" />
      </bpmndi:BPMNEdge>
      <bpmndi:BPMNEdge id="Flow_0ez4w93_di" bpmnElement="Flow_0ez4w93">
        <di:waypoint x="508" y="200" />
        <di:waypoint x="650" y="200" />
      </bpmndi:BPMNEdge>
      <bpmndi:BPMNEdge id="Association_0c0ii8c_di" bpmnElement="Association_0c0ii8c">
        <di:waypoint x="839" y="130" />
        <di:waypoint x="853" y="81" />
      </bpmndi:BPMNEdge>
      <bpmndi:BPMNShape id="TextAnnotation_1q9vfnx_di" bpmnElement="TextAnnotation_1q9vfnx">
        <dc:Bounds x="810" y="40" width="100" height="41" />
        <bpmndi:BPMNLabel />
      </bpmndi:BPMNShape>
      <bpmndi:BPMNEdge id="Flow_03o4pmp_di" bpmnElement="Flow_03o4pmp">
        <di:waypoint x="940" y="195" />
        <di:waypoint x="1102" y="195" />
      </bpmndi:BPMNEdge>
      <bpmndi:BPMNEdge id="Flow_040uk43_di" bpmnElement="Flow_040uk43">
        <di:waypoint x="188" y="195" />
        <di:waypoint x="250" y="195" />
      </bpmndi:BPMNEdge>
      <bpmndi:BPMNEdge id="Flow_0fmvfja_di" bpmnElement="Flow_0fmvfja">
        <di:waypoint x="350" y="195" />
        <di:waypoint x="440" y="195" />
      </bpmndi:BPMNEdge>
    </bpmndi:BPMNPlane>
  </bpmndi:BPMNDiagram>
</bpmn:definitions>
\end{verbatim}
}
\caption{BPMN code corresponding to the prior sequence of Figure \ref{fig:flowbench_test}}
\label{fig:prior_sequence}
\end{figure*}

\begin{figure*} [hp]
{\tiny
\begin{verbatim}
<?xml version="1.0" encoding="UTF-8"?>
<bpmn:definitions xmlns:xsi="http://www.w3.org/2001/XMLSchema-instance" xmlns:bpmn="http://www.omg.org/spec/BPMN/20100524/MODEL" 
xmlns:bpmndi="http://www.omg.org/spec/BPMN/20100524/DI" 
xmlns:dc="http://www.omg.org/spec/DD/20100524/DC" xmlns:di="http://www.omg.org/spec/DD/20100524/DI" exporter="Camunda Modeler" exporterVersion="5.32.0">
  <bpmn:process id="Process_1j6betq" isExecutable="false">
    <bpmn:startEvent id="StartEvent_1twgfyv">
      <bpmn:outgoing>Flow_040uk43</bpmn:outgoing>
    </bpmn:startEvent>
    <bpmn:subProcess id="Activity_0n3dkn6">
      <bpmn:incoming>Flow_0fmvfja</bpmn:incoming>
      <bpmn:outgoing>Flow_03o4pmp</bpmn:outgoing>
      <bpmn:multiInstanceLoopCharacteristics isSequential="true" />
      <bpmn:startEvent id="Event_1g6k28n">
        <bpmn:outgoing>Flow_0ez4w93</bpmn:outgoing>
      </bpmn:startEvent>
      <bpmn:endEvent id="Event_0lbdydr">
        <bpmn:incoming>Flow_05t21yg</bpmn:incoming>
      </bpmn:endEvent>
      <bpmn:sequenceFlow id="Flow_0ez4w93" sourceRef="Event_1g6k28n" targetRef="Activity_0sj4qjl" />
      <bpmn:task id="Activity_0sj4qjl" name="GitHub_Issue__3_0_0__create_Issue">
        <bpmn:incoming>Flow_0ez4w93</bpmn:incoming>
        <bpmn:outgoing>Flow_05t21yg</bpmn:outgoing>
      </bpmn:task>
      <bpmn:sequenceFlow id="Flow_05t21yg" sourceRef="Activity_0sj4qjl" targetRef="Event_0lbdydr" />
    </bpmn:subProcess>
    <bpmn:endEvent id="Event_1ycwwda">
      <bpmn:incoming>Flow_03o4pmp</bpmn:incoming>
    </bpmn:endEvent>
    <bpmn:sequenceFlow id="Flow_03o4pmp" sourceRef="Activity_0n3dkn6" targetRef="Event_1ycwwda" />
    <bpmn:task id="Activity_0cwpd7f" name="GitHub_Repository__3_0_0__retrievewithwhere_Repository">
      <bpmn:incoming>Flow_040uk43</bpmn:incoming>
      <bpmn:outgoing>Flow_0fmvfja</bpmn:outgoing>
    </bpmn:task>
    <bpmn:sequenceFlow id="Flow_040uk43" sourceRef="StartEvent_1twgfyv" targetRef="Activity_0cwpd7f" />
    <bpmn:sequenceFlow id="Flow_0fmvfja" sourceRef="Activity_0cwpd7f" targetRef="Activity_0n3dkn6" />
    <bpmn:textAnnotation id="TextAnnotation_1q9vfnx">
      <bpmn:text>for repo in repositories</bpmn:text>
    </bpmn:textAnnotation>
    <bpmn:association id="Association_0c0ii8c" associationDirection="None" sourceRef="Activity_0n3dkn6" targetRef="TextAnnotation_1q9vfnx" />
  </bpmn:process>
  <bpmndi:BPMNDiagram id="BPMNDiagram_1">
    <bpmndi:BPMNPlane id="BPMNPlane_1" bpmnElement="Process_1j6betq">
      <bpmndi:BPMNShape id="_BPMNShape_StartEvent_2" bpmnElement="StartEvent_1twgfyv">
        <dc:Bounds x="152" y="177" width="36" height="36" />
      </bpmndi:BPMNShape>
      <bpmndi:BPMNShape id="Activity_0cwpd7f_di" bpmnElement="Activity_0cwpd7f">
        <dc:Bounds x="250" y="155" width="100" height="80" />
        <bpmndi:BPMNLabel />
      </bpmndi:BPMNShape>
      <bpmndi:BPMNShape id="Event_1ycwwda_di" bpmnElement="Event_1ycwwda">
        <dc:Bounds x="1102" y="177" width="36" height="36" />
      </bpmndi:BPMNShape>
      <bpmndi:BPMNShape id="Activity_0dishkm_di" bpmnElement="Activity_0n3dkn6" isExpanded="true">
        <dc:Bounds x="440" y="130" width="500" height="150" />
      </bpmndi:BPMNShape>
      <bpmndi:BPMNShape id="Event_1g6k28n_di" bpmnElement="Event_1g6k28n">
        <dc:Bounds x="472" y="182" width="36" height="36" />
      </bpmndi:BPMNShape>
      <bpmndi:BPMNShape id="Activity_0sj4qjl_di" bpmnElement="Activity_0sj4qjl">
        <dc:Bounds x="650" y="160" width="100" height="80" />
        <bpmndi:BPMNLabel />
      </bpmndi:BPMNShape>
      <bpmndi:BPMNShape id="Event_0lbdydr_di" bpmnElement="Event_0lbdydr">
        <dc:Bounds x="842" y="182" width="36" height="36" />
      </bpmndi:BPMNShape>
      <bpmndi:BPMNEdge id="Flow_05t21yg_di" bpmnElement="Flow_05t21yg">
        <di:waypoint x="750" y="200" />
        <di:waypoint x="842" y="200" />
      </bpmndi:BPMNEdge>
      <bpmndi:BPMNEdge id="Flow_0ez4w93_di" bpmnElement="Flow_0ez4w93">
        <di:waypoint x="508" y="200" />
        <di:waypoint x="650" y="200" />
      </bpmndi:BPMNEdge>
      <bpmndi:BPMNEdge id="Association_0c0ii8c_di" bpmnElement="Association_0c0ii8c">
        <di:waypoint x="839" y="130" />
        <di:waypoint x="853" y="81" />
      </bpmndi:BPMNEdge>
      <bpmndi:BPMNShape id="TextAnnotation_1q9vfnx_di" bpmnElement="TextAnnotation_1q9vfnx">
        <dc:Bounds x="810" y="40" width="100" height="41" />
        <bpmndi:BPMNLabel />
      </bpmndi:BPMNShape>
      <bpmndi:BPMNEdge id="Flow_03o4pmp_di" bpmnElement="Flow_03o4pmp">
        <di:waypoint x="940" y="195" />
        <di:waypoint x="1102" y="195" />
      </bpmndi:BPMNEdge>
      <bpmndi:BPMNEdge id="Flow_040uk43_di" bpmnElement="Flow_040uk43">
        <di:waypoint x="188" y="195" />
        <di:waypoint x="250" y="195" />
      </bpmndi:BPMNEdge>
      <bpmndi:BPMNEdge id="Flow_0fmvfja_di" bpmnElement="Flow_0fmvfja">
        <di:waypoint x="350" y="195" />
        <di:waypoint x="440" y="195" />
      </bpmndi:BPMNEdge>
    </bpmndi:BPMNPlane>
  </bpmndi:BPMNDiagram>
</bpmn:definitions>

\end{verbatim}
}
\caption{BPMN code corresponding to the final output of Figure \ref{fig:flowbench_test}}
\label{fig:final_output}
\end{figure*}